\documentclass[journal]{IEEEtran}

\usepackage[dvips]{graphicx}

\usepackage{amsmath}
\usepackage{amssymb}
\usepackage{subfigure}
\usepackage{multirow}
\usepackage[colorlinks, linkcolor=red, citecolor=green]{hyperref}

\usepackage{makecell}
\usepackage{tabularx}
\usepackage[numbers,sort]{natbib}

\begin{document}
%
\title{Adversarial Generation of Training Examples: Applications to Moving Vehicle License Plate Recognition}
%
%
%
	
\author{Xinlong~Wang, Zhipeng~Man, Mingyu~You,~Chunhua~Shen
\thanks{X. Wang, Z. Man, M. You  are with  College of Electronic and Information Engineering, Tongji University, Shanghai China (e-mail: wangxinlon@gmail.com).}
\thanks{C. Shen is with School of Computer Science, The University of Adelaide, SA 5005, Australia (e-mail: chunhua.shen@adelaide.edu.au).}
}

\maketitle

\begin{abstract}
	Generative Adversarial Networks (GAN) have attracted much research attention
	recently,  leading to impressive results for natural image generation.
	However, to date little success was observed in using GAN generated images for
	improving classification tasks.
	Here we attempt to explore, in the context of  license plate recognition for moving cars using a moving camera,
	whether it is possible to generate
	synthetic training data using GAN to improve recognition accuracy.
	With a carefully-designed pipeline, we show that the answer is affirmative.
	First, a large-scale image set is generated using the generator of GAN,
	without manual annotation. Then, these images are fed to a deep
	convolutional neural network (DCNN) followed by a bidirectional recurrent
	neural network (BRNN) with long short-term memory (LSTM), which performs
	the feature learning and sequence labelling. Finally, the pre-trained
	model is fine-tuned on real images.

	Our experimental results on a few data sets demonstrate the effectiveness
	of using GAN images: an improvement of 7.5 recognition accuracy percent points (pp) over a strong
	baseline with moderate-sized real data being available.
	We show that the proposed framework achieves competitive recognition
	accuracy on challenging test datasets. We also leverage the depthwise separate convolution   	to construct a lightweight convolutional recurrent neural network (LightCRNN), which is about half size and 2$\times$ faster on CPU. Combining this framework and the proposed pipeline, we make progress in performing accurate recognition on mobile and embedded devices.
\end{abstract}


%
\IEEEpeerreviewmaketitle

\section{Introduction}

\IEEEPARstart{A}{s} GANs \cite{goodfellow2014generative, radford2015unsupervised} have been developed to generate compelling natural images, we attempt to
explore whether GAN-generated images can help training as well as real images.
GANs are used to generate images
through an adversarial training procedure that learns the real
data distribution. Most of previous work focuses on  using GANs to manipulate images
for computer graphics applications
\cite{isola2016image,reed2016generative}.
However, to date little success was observed in using GAN generated images for
improving classification tasks.
To our knowledge, we are one of the first to exploit GAN-generated images to supervised recognition tasks
and gain remarkable improvements.
For many tasks, labelled data are limited and
hard to collect, not to mention the fact that training data with manual annotation
typically costs tremendous effort.  This problem becomes worse in the deep learning era as
in general deep learning methods are much more data-hungry.
Applications are   hampered   in these fields
when training data for the  task is scarce. Inspired by the excellent
ph\-o\-to-real\-is\-tic quality of images generated by GANs, we explore its effectiveness in
supervised learning tasks.

More specifically,  in this work we study the problem  of using GAN images in the context of
license
plate recognition (LPR). License plate recognition is an important research
topic, as a crucial part in intelligent transportation systems (ITS). A
license plate is a principal identifier of a vehicle. This binding
relationship creates many applications using license plate recognition,
such as traffic control, electronic payment systems and criminal
investigation.
{\em
Most research on LPR  have been focused on using a fixed camera to capture car license images.
}
However,  the task of using a moving camera to capture license plate images of moving cars
can be much more challenging. This task is not well studied in the literature. Moreover,
Lack of a large labelled training data for this task renders data-hungry deep learning methods to not well.

Recent advances in deep learning algorithms bring multiple
choices which can achieve great performance in specific datasets. A notable
cascade framework \cite{li2016reading} applies a sliding window to feature
extraction, which perform the LPR without segmentation. Inspired by the
work of  \cite{shi2016end,li2016reading} and CNN's strong capabilities of
learning informative representations from images, we combine deep
convolutional neural network and recurrent neural network in a single framework,
and recognize a license plate without segmentation, which views a
license plate as a whole object and learns its inner rules from
data. Independent of the deep models being
applied, to achieve a high recognition accuracy requires a large amount of annotated
license plate image data.
However, due to privacy issues, image data of license plates are hard
to collect. Moreover, manual annotation is expensive and time-consuming. What is
more, LPR is a task with regional characteristic: license plates differ in
countries and regions.   For many of regions, it is even harder to collect
sufficient samples to build a robust and high-performance model.
To tackle this problem, we
propose a novel pipeline which leverages GAN-generated images in training. The
experiments on Chinese license plates show its effectiveness especially
when training data are scarce.

To achieve this goal,
first, we use computer graphic scripts to synthesize
images of license plates, following the fonts, colors and composition rules.
Note that these synthetic images are not photo realistic, as
{\em it is very difficult to model
the complex distribution of real license plate images by using hand-crafted rules only.}
This step is necessary as we use these synthetic images as the input to the GAN.
Thus the GAN model refines the image style to make images look close to real images.
At the same time the content, which is the actual vehicle plate number, is kept.
This is important as we need the plate number---the ground-truth label---for supervised classification in the sequel.
We then
train a CycleGAN \cite{zhu2017unpaired} model to generate realistic images of
license plates with labels, by learning the mapping from synthetic images to real
images.
The main reason for using CycleGAN is that, we do not have
paired input-output  examples for training the GAN.
As such, methods like pixel-to-pixel GAN \cite{isola2016image} are not applicable.
CycleGAN
can learn to translate between domains without paired input-output examples.
An assumption of CycleGAN is that  there is an underlying relationship
between the two domains. In our case, this relationship is
two different renderings of the same license plate:  synthetic graphics versus
real photos. {\em We rely on CycleGAN  to learn the translation from
synthetic graphics to real photos}.

\begin{figure*}[t!]
  \centering
  \subfigure[The CycleGAN Model]{
    \label{pipeline:a}
    \includegraphics[width=2.75in]{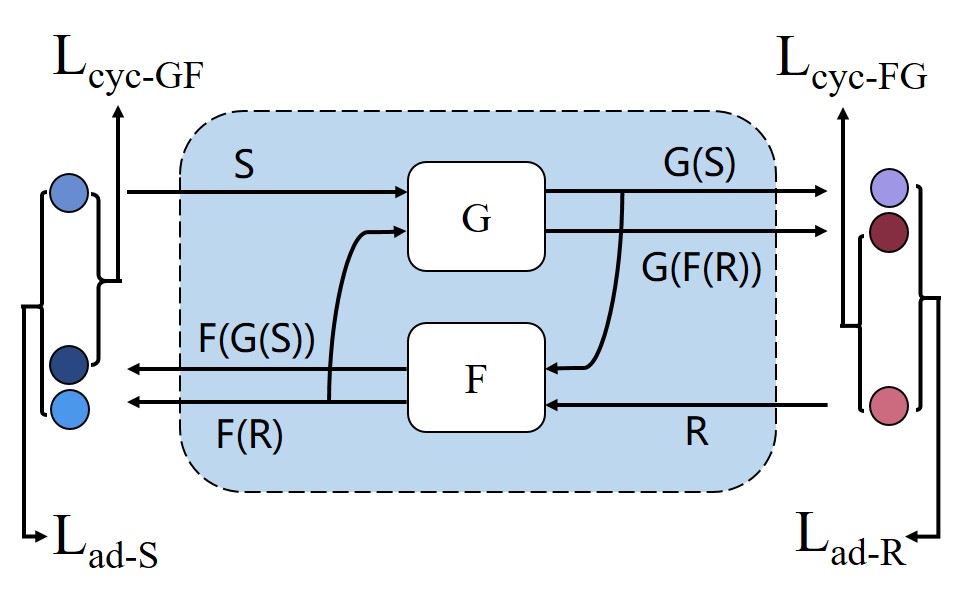}}
  \hspace{.1in}
  \subfigure[Our proposed Pipeline]{
    \label{pipeline:b} %
    \includegraphics[width=2.5in]{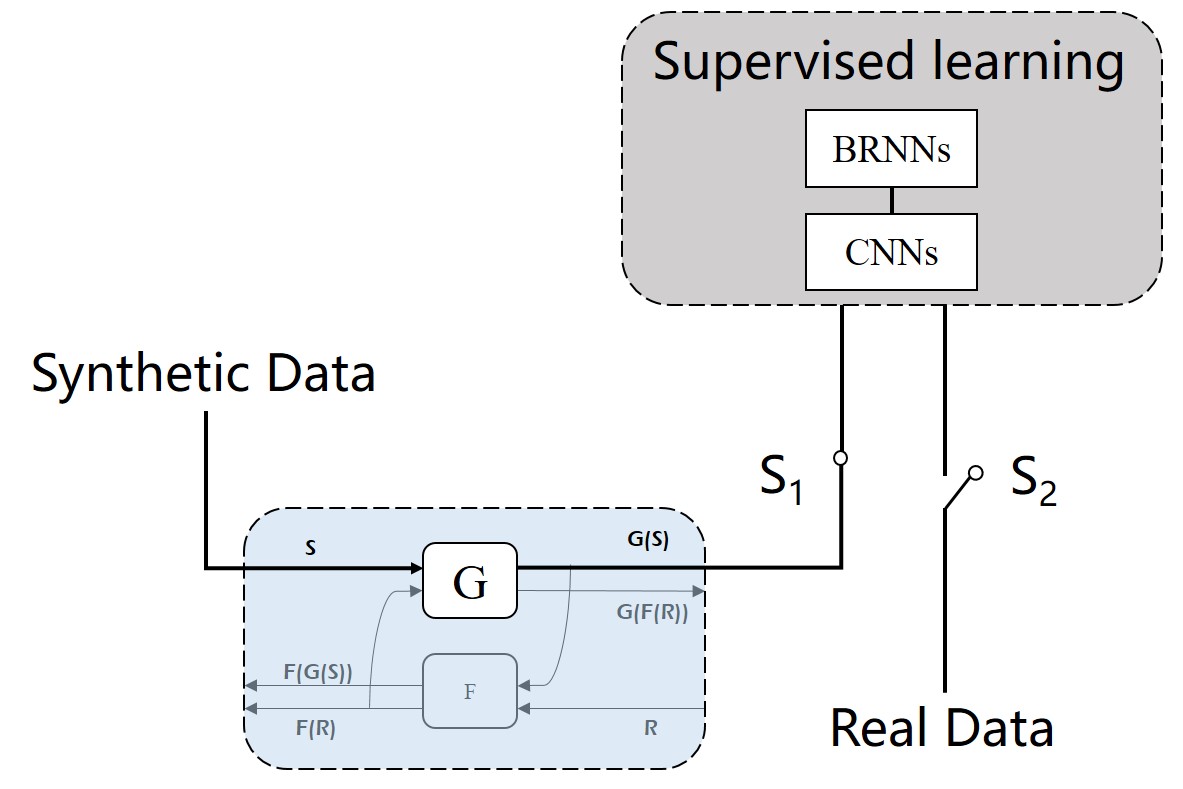}}
  \caption{(a) The architecture of the CycleGAN model, which is essential for our framework.
    There are two generators
    (or mapping functions): $G$ and $F$. $S$ and $R$ represent the synthetic and real images respectively. Adversarial loss ($L_{\rm ad}$) is used
    to evaluate the performance of each generator and corresponding
    discriminator. Besides, the cycle consistency loss ($L_{\rm cyc}$) is employed
    to evaluate whether $G$ and $F$ are cycle-consistent. (b) The pipeline of
    proposed approach. There are two components: a generative adversarial model
    and a convolutional recurrent neural network. The ``Synthetic Data''
    represent the labeled images generated by our scripts.
}
  \label{pipeline} %
\end{figure*}

Finally, we train the convolutional recurrent neural network
\cite{shi2016end} on the GAN images, and fine-tune it on real images, as
illustrated in Fig. \ref{pipeline:b} in the manner of curriculum learning
paradigm \cite{BengioCL}.

The core insight behind this is that we exploit the license number from
synthetic license plates and the photo-realistic style from real license
plates, which combines knowledge-based (i.e., hand-crafted rules) and learning-based
data-driven
approaches.
Applying the proposed pipeline, we achieve significantly improved recognition accuracy
on moderate-sized training sets of real images. Specifically, the model trained on 25\% training data (50$k$) achieves an impressive accuracy of 96.3\%, while the accuracy is 96.1\% when
trained on the whole training set (200$k$) without our approach.

Besides the proposed framework, we consider semi-supervised learning using
unlabeled images generated by deep convolutional GAN (DCGAN)
\cite{radford2015unsupervised} as an add-on, for the sake of comparison between
our approach and existing semi-supervised method using GAN images.

To run the deep LPR models on mobile devices, we replace the standard convolutions with depthwise separate convolutions \cite{sifre2014rigid} and construct a lightweight convolutional RNN. This architecture efficiently decreases the model size and speed up the inference process, with only a small reduction in accuracy.

The main contributions of this work can be summarized as follows.
\begin{itemize}
\item We propose a pipeline that applies GAN-generated images to supervised learning as auxiliary data.
Furthermore,
    we follow a curriculum learning strategy  to train the system with gradually more complex training data.
    In the first step, we use a large number of  GAN-generated images whose appearance  is simpler than real images,
        and a large lexicon of license plate numbers\footnote{Recall that the vehicle license plate numbers/letters follow
        some rules or grammars. Our system learns the combinatorial rules with BRNN on the large number of GAN images.}.
        At the same time, the first step finds  a good initialization of the appearance model for the second step.
        By employing relatively small real-world images,
        the system gradually learns how to handle complex appearance patterns.

\item We show the effectiveness of Wasserstein distance loss in CycleGAN training,
      which aids data variety and converged much better.
      The original CycleGAN uses  a least-squares loss for the adversarial loss, which is more stable
      than the standard maximum-likelihood loss.
        We show that the Wasserstein distance loss as in Wasserstein GAN (WGAN) \cite{arjovsky2017wasserstein}
        works better for CycleGAN as well.

\item We show that when and why GAN images help in supervised learning.

\item We build a small and efficient LightCRNN, which is half model size and 2$\times$ faster on CPU.

\end{itemize}

The rest of paper is arranged as follows. In
Section 2 we review the related works briefly. In Section 3 we describe the details of networks used in our approach.
Experimental results are provided in Section 4, and conclusions are drawn in Section 5.

\section{Related Work}
In this section, we discuss the relevant work on license plate recognition, generative adversarial networks and data generation for training.

\subsection{License Plate Recognition}

    Existing methods on license plate recognition (LPR) can be divided into two
    categories: segmentation-based
    \cite{nomura2005novel,gou2016vehicle,guo2008license}
    and segmentation free-based \cite{li2016reading}. Segmentation-based
    methods firstly segment the license plate into individual characters,
    and then recognize the segmented
    character respectively using a classifier.
    Segmentation algorithms mainly consist of projection-based
    \cite{nomura2005novel,guo2008license} and connected component-based
    \cite{anagnostopoulos2006license,jiao2009configurable}. After the
    segmentation, template matching based
    \cite{rasheed2012automated,goel2013vehicle} and learning based
    \cite{wen2011algorithm,llorens2005car,jiao2009configurable} algorithms can
    be used to tackle this character level classification task.

    Learning based
    algorithms including support vector machine \cite{wen2011algorithm}, hidden
    Markov model (HMM) \cite{llorens2005car} and  neural networks
    \cite{jiao2009configurable,tao2016principal} are more robust than the template matching based methods since they extract
    discriminative features. However, the segmentation process loses
    information about inner rules in license plates and the segmentation
    performance has a significant influence on the recognition performance. Li
    and Shen \cite{li2016reading} proposed a cascade framework using deep
    convolutional neural networks and LSTMs for license plate recognition
    without segmentation, where a sliding window is applied to extracting
    feature sequence.
    Our method is also a segmentation-free approach based on
    framework proposed by \cite{shi2016end}, where a deep CNN is applied for
    feature extracting directly without a sliding window, and a bidirectional
    LSTM network is used for sequence labeling.

\subsection{Generative Adversarial Networks}

    The generative adversarial networks \cite{goodfellow2014generative} train a generative model and discriminative
    model simultaneously via an adversarial process. Deep convolutional generative adversarial networks (DCGANs)
    \cite{radford2015unsupervised} provide a stable architecture for training GANs. Conditional GAN
    \cite{mirza2014conditional} generate images with specific class labels by conditioning on both the generator and
    discriminator. Not only class labels, GANs can be conditioned on text descriptions \cite{reed2016generative} and
    images \cite{isola2016image}, which build a text-to-image or image-to-image translation.

    Zhu et al.\ \cite{zhu2017unpaired} proposed the CycleGAN.
    It learns a mapping between two domains without paired images, upon which our model
    builds.
    In order to use unpaired images for training, CycleGAN introduces the cycle consistency loss
    to fulfill the idea of ``If we translate from one domain to another and back again we
    must arrive where we start".

    The work of Wasserstein GAN (WGAN) \cite{arjovsky2017wasserstein} designed a
    training algorithm that provides some techniques to improve the stability
    of learning and prevent from mode collapse. Beyond that,  GANs have also
    achieved impressive results in image inpainting \cite{pathak2016context},
    representation learning \cite{salimans2016improved} and 3D object generation
    \cite{3dgan}.

However, to date, little results were reported to demonstrate the GAN images'
effectiveness in supervised learning. We propose to use CycleGAN \cite{zhu2017unpaired}
and techniques in WGAN to generate labeled images and show that these generated
images indeed help improve the performance of recognition.

\subsection{Data Generation for Training}

    Typically, supervised deep networks achieve acceptable performance only with a large
    amount of labeled examples available.
    The performance often improves as the dataset
    size increases. However, in many cases, it is hard or even impossible
    to obtain a large number of labeled data.

    Synthetic data are  used to show great performance in text
    localisation \cite{gupta2016synthetic} and scene text recognition
    \cite{jaderberg2014synthetic}, without manual annotation. Additional
    synthetic training data \cite{lim2017evolutionary} yields improvements in person detection
    \cite{yu2010improving}, font recognition \cite{wang2015deepfont} and semantic segmentation \cite{ros2016synthia}. But, the knowledge-based approach which hard encodes
    knowledge about what the real images look like are fragile,  as the generated
    examples often look like toys to discriminative model when compared to real
    images. Zheng et al. \cite{zheng2017unlabeled} use unlabeled samples
    generated by a vanilla DCGAN for semi-supervised learning, which slightly improves the
    person re-identification performance. In this work, we combine
    knowledge-based approach and learning-based approach, to generate labeled
    license plates from generator of GANs for supervised training. For
    comparison, we also perform a semi-supervised learning using unlabeled GAN
    images.

\section{Network Overview}

    In this section, the pipeline of the proposed method is described. As shown in Fig. \ref{pipeline}, we train the GAN
    model using synthetic images and real images simultaneously. We then use the generated images to train a
    convolutional BRNN model.
    Finally we fine tune this model using real images. Below is an illustration of the two components, namely, the GAN and
    the convolutional RNN, in detail.

\subsection{Generative Adversarial Network}

    The generative adversarial networks generally consists of two sub-networks: a generator and a discriminator. The
    adversarial training process is a minimax game: both sub-networks aim to minimize its own cost and maximize the
    other's cost. This adversarial process leads to a  converged  status that the generator outputs realistic images and
    the discriminator extracts deep features. The GAN frameworks we used in this pipeline are Cycle-Consistent
    Adversarial Networks (CycleGAN) \cite{zhu2017unpaired} with  the Wasserstein distance loss as in
    Wasserstein GAN (WGAN) \cite{arjovsky2017wasserstein}.

\def\cS{ {\cal S}}
\def\cR{ {\cal R}}

\textbf{CycleGAN}
As our goal is to learn a mapping that maps synthetic images into real images, namely domain ${\cal S}$ to
domain ${\cal R}$, we use two generators: $G : \cS \rightarrow \cR$, $F : \cR \rightarrow \cS$,
and two discriminators $D_S$ and $D_R$.
The core insight behind this is that, we not only want to have $G(S) \approx R$, but also $F(G(S)) \approx S$, which is an
additional restriction used to prevent the well-known problem of mode collapse, where all input images map into the same
output image. As illustrated in Fig. \ref{pipeline:a}, mapping functions $G$ and $F$ both should be cycle-consistent. We
follow the settings in \cite{zhu2017unpaired}. The generator network consists of two stride-2 convolutions, six residual
blocks and two fractionally-strided convolutions with stride $\frac{1}{2}$, i.e., deconvolutions that upsample
the feature maps.
For the discriminator networks, we use $70
\times 70$ PatchGANs \cite{isola2016image}, which, instead of classifying the entire image,
classifies $70 \times 70$ image patches to be
fake or real. The objective is defined as the sum of adversarial loss and cycle consistency loss: \begin{equation}
    \begin{split} L(G, F, D_S, D_R) = &L_{\rm LSGAN}(G, D_R, S, R)\\ &+ L_{\rm LSGAN}(F, D_S, R, S) \\  &+ \lambda
    L_{\rm cyc}(G, F). \end{split} \end{equation}
Here $L_{LSGAN}(G, D_R, S, R)$ evaluates
    the adversarial least-squares loss of the mapping function $G:S \rightarrow R$ and discriminator $D_R$:
\begin{equation}
\begin{split}
L_{\rm LSGAN}(G, D_R, S, R) = &E_{r \sim p_{data}(r)}[(D_R(r) - 1)^2] \\ &+ E_{s \sim p_{data}(s)}[D_R(G(s))^2].
\end{split}
\end{equation}
$G$ tries to minimize (2) against $D_R$ that tries to maximize it: while $G$ aims to generate images $G(s)$ that look similar to real images, $D_R$ tries to distinguish between $G(s)$ and real images $r$, and vice versa. Cycle consistency loss is defined as below:
\begin{equation}
\begin{split}
L_{\rm cyc}(G, F) = & E_{s \sim p_{data}(s)}[{\Vert F(G(s)) - s\Vert}_1]\\ &+ E_{r \sim p_{data}(r)}[{\Vert G(F(r)) - r\Vert}_1].
\end{split}
\end{equation}
The optimization of $(3)$ brings $F(G(s))$ back to $s$, $G(F(r))$ to $r$. $\lambda$ in (1) represent the strength of the cycle consistency. Details of CycleGAN can be referred to \cite{zhu2017unpaired}.

\textbf{Wasserstein GAN}
We apply the techniques proposed in \cite{arjovsky2017wasserstein} to CycleGAN and propose the CycleWGAN. In training of GANs, the method of
    measuring the distance between two distances plays a crucial role. A poor evaluation of distance makes it hard to
    train, even causes mode collapse. The Earth-Mover (EM) distance, a.k.a.\ Wasserstein distance,
    is used when deriving the loss metric, as it is more sensible and converges better. The EM distance between two distributions $P_r$ and $P_s$ is defined as below:
\begin{equation}
EM(P_r,P_s) = \inf_{\lambda \in \prod (P_r, P_s)} E_{(x,y)\sim \lambda} [\Vert x - y \Vert],
\end{equation}
which can be interpreted as the ``mass'' must be moved to transform the distribution $P_r$ into the distribution $P_s$.
To make it tractable, we apply the Kantorovich-Rubinstein duality and get
\begin{equation}
EM(P_r,P_s) = \sup_{{\Vert f \Vert}_L \leq 1} E_{x \sim P_r}[f(x)] - E_{x \sim P_s}[f(x)],
\end{equation}
where $f$ are 1-Lipschitz function.
To solve the problem
\begin{equation}
\max_{{\Vert f \Vert}_L \leq 1} E_{x \sim P_r}[f(x)] - E_{x \sim P_s}[f(x)],
\end{equation}
we can find a parameterized family of functions ${\lbrace f_w \rbrace}_{w \in W}$ which are $K$-Lipschitz for some constant $K$ such that ${\Vert f \Vert}_L \leq K$. Then this problem is transformed to
\begin{equation}
\max_{w \in W} E_{x \sim P_r}[f(x)] - E_{x \sim P_s}[f(x)].
\end{equation}
In order to enforce a Lipschitz constraint that parameters $w$ range in a compact space, we clamp the weights to a fixed range after each gradient update. In the training of GAN, the discriminator play the role of $f$ above.

Specifically, we make four modifications based on CycleGAN described above:
\begin{itemize}
 \item Replace the original adversarial loss (2) with:
 \begin{equation}
 \begin{split} L_{\rm WGAN}(G, D_R, S, R) = &E_{r \sim p_{data}(r)}[D_R(r)] \\ &- E_{s \sim p_{data}(s)}[D_R(G(s))].
 \end{split}
 \end{equation} The same to $L_{\rm LSGAN}(F, D_S, R, S)$.
 \item Update discriminators $d_{iter}$ times before each update of generators.
 \item Clamp the parameters of discriminators to a fixed range after each gradient update.
 \item Use RMSProp \cite{Tieleman2012} for optimization, instead of Adam \cite{kingma2014adam}.
 \end{itemize}

\begin{table}[t!]
\footnotesize
\centering
\caption{Configuration of the convolutional RNN model. ``k'', ``s'' and ``p'' represent kernel size, stride and padding size.}
\label{crnn_config}
\begin{tabular}{c|c}
\hline
\hline
Layer Type & Configurations \\
\hline
\hline
Bidirectional-LSTM & \#hidden units: 256\\
\hline
Bidirectional-LSTM & \#hidden units: 256\\
\hline
ReLu & - \\
\hline
BatchNormalization & - \\
\hline
Convolution & \#filters:512, k:$2 \times 2$, s:1, p:0 \\
\hline
ReLu & - \\
\hline
BatchNormalization & - \\
\hline
Convolution & \#filters:512, k:$2 \times 2$, s:1, p:0 \\
\hline
MaxPooling & s:$1 \times 2$, p:$1 \times 0$ \\
\hline
ReLu & - \\
\hline
Convolution & \#filters:512, k:$3 \times 3$, s:1, p:1 \\
\hline
ReLu & - \\
\hline
BatchNormalization & - \\
\hline
Convolution & \#filters:512, k:$3 \times 3$, s:1, p:1 \\
\hline
MaxPooling & s:$1 \times 2$, p:$1 \times 0$ \\
\hline
ReLu & - \\
\hline
Convolution & \#filters:256, k:$3 \times 3$, s:1, p:1 \\
\hline
ReLu & - \\
\hline
BatchNormalization & - \\
\hline
Convolution & \#filters:256, k:$3 \times 3$, s:1, p:1 \\
\hline
MaxPooling & p:$2 \times 2$, s:2 \\
\hline
ReLu & - \\
\hline
Convolution & \#filters:128, k:$3 \times 3$, s:1, p:1 \\
\hline
MaxPooling & p:$2 \times 2$, s:2 \\
\hline
ReLu & - \\
\hline
Convolution & \#filters:64, k:$3 \times 3$, s:1, p:1 \\
\hline
Input & $160 \times 48 \times 3$ RGB images\\
\hline
\hline
\end{tabular}
\end{table}

\begin{table*}[h!]
\footnotesize
\centering
\caption{Composition of the real datasets. The two datasets are collected separately but both come from open environments.}
\label{datasets}
\begin{tabular}{cccccccc}
\hline
Name & Width & Height & Time & Style & Provinces & Training set size & Test set size \\
\hline
Dataset-1 & $100\sim160$ & $30\sim50$ & Day, Night & Blue, Yellow & 30/31 &203,774 & 9,986  \\
Dataset-2 & $50\sim150$  & $15\sim40$ & Day, Night & Blue, Yellow & 4/31 &45,139   & 5,925  \\
\hline
\end{tabular}
\end{table*}

\subsection{Convolutional Recurrent Neural Network}

The framework that we use is a combination of deep convolutional
neural networks and recurrent neural networks based on \cite{shi2016end}.
The recognition procedure consists of three
components: Sequence feature learning, sequence labelling and sequence decoding, corresponding to the convolutional RNN
architecture of convolutional layers, recurrent layers and decoding layer one to one. The overall configurations are
presented in Table \ref{crnn_config}.

\textbf{Sequence feature   learning}
CNNs have demonstrated impressive ability
of extracting deep features from image \cite{wei2017cross}. An 8-layer CNN model is applied to
extracting sequence features, as shown in Table \ref{crnn_config}. Batch
normalization is applied in 3rd, 5th, 7th and 8th layers, and rectified linear
units are followed after all these 8 layers. All three channel in RGB images
are used: images are resized to $160 \times 48 \times 3$ before fed to
networks. After the 8-layer extraction, a sequence of feature vectors is
outputted, as the informative representation of the raw images, and the input
for the recurrent layers.

\begin{figure}[t!]
\centering
\includegraphics[width=2.5in]{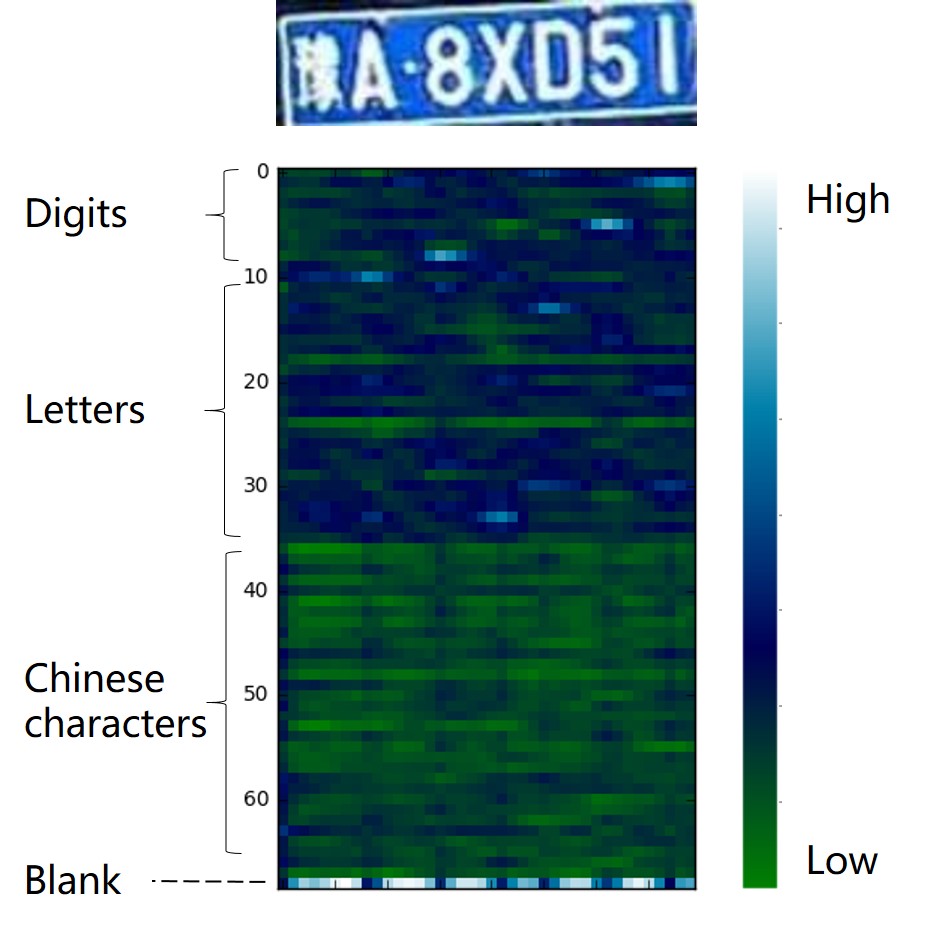}
\caption{License plate recognition confidence map. The top is the license plate to be predicted. The color map is the recognition probabilities from the linear projections of output by the last LSTM layer. The recognition probabilities on 68 classes are shown vertically (with classes order from top to bottom: 0-9, A-Z, Chinese characters and ``blank''). Confidence increases from green to white.}
\label{output_map}
\end{figure}

\textbf{Sequence labelling} The RNNs have achieved great success in sequential
problems such as handwritten recognition and language translation. To learn
contextual cues in license numbers, a deep bidirectional recurrent neural
network is built on the top of CNNs. To avoid gradient vanishing, LSTM is
employed, instead of vanilla RNN unit. LSTM is a special type of RNN unit,
which consists of a memory cell and gates. The form of LSTM in
\cite{zaremba2014recurrent} is adopted in this paper, where a LSTM contains a
memory cell and four multiplicative gates, namely input, input modulation,
forget and output gates. A fully-connected layer with 68 neurons is added
behind the last BiLSTM layer, for the 68 classes of label---31 Chinese
characters, 26 letters, 10 digits and ``blank''. Finally the feature sequence
is transformed into a sequence of probability distributions ${\bf X} = \{x^1, x^2,
..., x^T\}$ where $T$ is the sequence length of input feature vectors and the
superscript can be interpreted as time step. $x^t$ is a probability
distribution of time $t$: $x^t_k$ is interpreted as the probability of
observing label $k$ at time $t$, which is a probability distribution over the
set $\mathcal{L}(T) $ where $T$ is the length and $\mathcal{L}$ contains all of
the 68 labels, as illustrated in Fig. \ref{output_map}.

\textbf{Sequence decoding}
Once we have
the sequence of probability distributions, we transform it into character string using best path decoding \cite{graves2006connectionist}.
The probability of a specific path $\pi$ given an input $x$ is defined as below:
\begin{equation}
p(\pi \mid x) = \prod_{t=1}^T y_{\pi_t}^t.
\end{equation}
As described above, $y_{\pi_t}^t$ means the probability of observing label $\pi_t$ at time $t$, which is visualized in Fig. \ref{output_map}.
For the best path decoding, the output is the labelling of most probable path:
\begin{equation}
l = \mathcal{B}({\pi}^*)\mathrm{,}
\end{equation}
$$\mathrm{where }\ {\pi}^* = arg \max_{\pi}p(\pi \mid x).$$
 Here we define the operation $\mathcal{B}$ as removing all blanks and repeated labels from the path. For example, $\mathcal{B}(aa-ab--) = \mathcal{B}(-a-aa-b) = aab$. As there is no tractable globally optimal decoding algorithm for this problem, the best path decoding is an approximate method that assumes the most probable path corresponds to the most probable labelling, which simply takes the most active output at each time step as $\pi^*$ and then map $\pi^*$ onto $l$ .
More sophisticated decoding methods such as dynamic programming may lead to better results.

\textbf{Training method}
We train the networks with stochastic gradient descent (SGD).
The labelling loss is derived using Connectionist Temporal Classification (CTC)
\cite{graves2006connectionist}. Then the optimization algorithm Adadelta \cite{zeiler2012adadelta} is applied,
as it converges fast and requires no manual setting of a learning rate.

\begin{figure}[t!]
\centering
\includegraphics[width=2.5in]{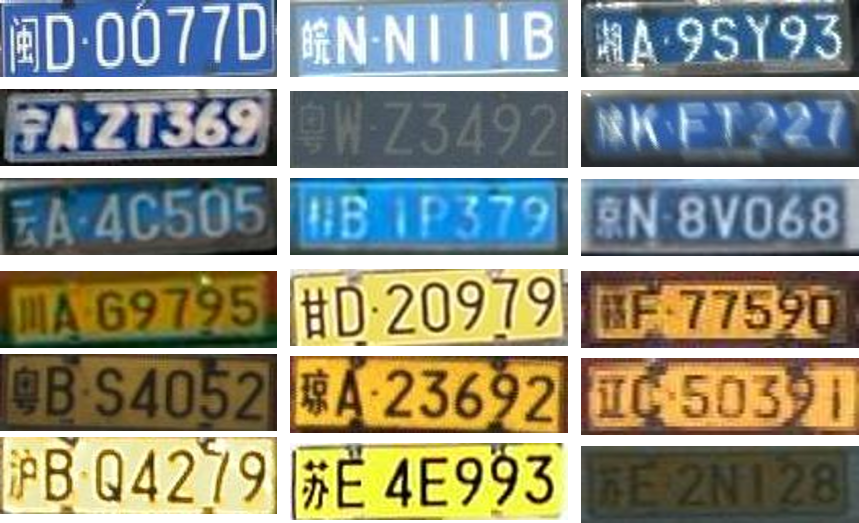}
\caption{Some Samples of Dataset-1. The license plates in Dataset-1 come from 30 different provinces of all 31 provinces. All of them are captured from real traffic monitoring scenes.}
\label{samples_Dataset_1}
\end{figure}

\begin{figure}[t]
\centering
\includegraphics[width=2.5in]{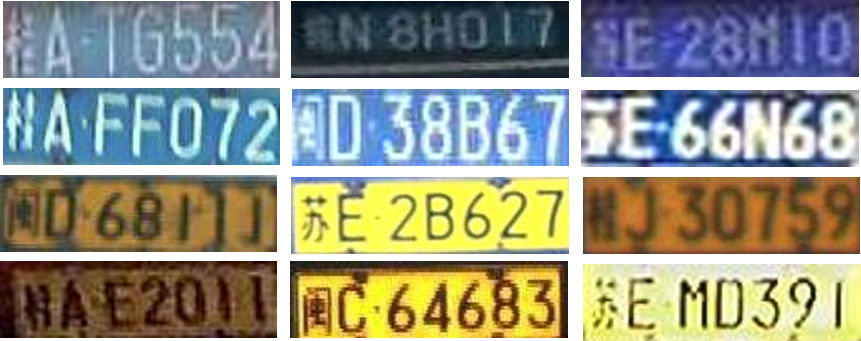}
\caption{Some samples of Dataset-2. All of license plates in Dataset-2 are collected from four provinces, from different channels with Dataset-1.}
\label{samples_Dataset_2}
\end{figure}

\section{Experiments}

\subsection{Datasets}
We collect two datasets in which images are captured from a wide variety of real traffic monitoring scenes under various viewpoints, blurring and illumination. As listed in Table \ref{datasets}, Dataset-1 contains a training set of 203,774 plates and a test set of 9,986 plates; Dataset-2 contains a training set of 45,139 plates and a test set of 5,925 plates. Some sample images are shown in Fig. \ref{samples_Dataset_1} and Fig. \ref{samples_Dataset_2}. For Chinese license plates, the first character is a Chinese character that represents province. While there are 31 abbreviations for all of the provinces, Dataset-1 contains 30 classes of them and Dataset-2 contains 4 classes of them, which means the source areas of our data covers a wide range of ground-truth license plates. Our datasets are very challenging and comprehensive due to the extreme diversity of character patterns, such as different resolution, illumination and character distortion. Note that these two datasets come from different sources and do not follow the same distribution.

\subsection{Evaluation Criterion} In this work, we evaluate the model performance in terms of
recognition accuracy and
character recognition accuracy, which are evaluated  at the license plate level and character level,
respectively. Recognition
accuracy is defined as the number of license plates that each character is correctly recognized divided by the total
number of license plates: \begin{equation} \mathrm{RA = \frac{Number\ of\ correctly\ recognized\ license\
plates}{Number\ of\ all\ license\ plates}}. \end{equation} Character recognition accuracy is defined as the number of
correctly recognized characters divided by the number of all characters: \begin{equation} \mathrm{CRA = \frac{Number\
    of\ correctly\ recognized\ characters}{Number\ of\ all\ characters}}. \end{equation}

 Besides, we compute the top-$N$
    recognition accuracy as well. Top-$N$ recognition accuracy is defined as the number of license plates
    that ground-truth label is in top $N$ predictions divided by the total number of license plates. The recognition
    accuracy defined above is a special case of the top-$N$ recognition with $N$ equals 1.
    Benefiting from LSTMs’ sequence
    output, we can easily  obtain  the top-$N$ recognition accuracy by making a beam search decoding on the output logits. The
    top-$N$ recognition accuracy make sense in criminal investigation: an accurate list of candidates can be provided
    when searching for a specific license plate, which means a high top-$N$ accuracy promise a low missing rate.

\begin{figure*}[htbp]
  \centering
  \subfigure[Script Images]{
    \label{gan_imgs:a} %
    \includegraphics[width=2.5in]{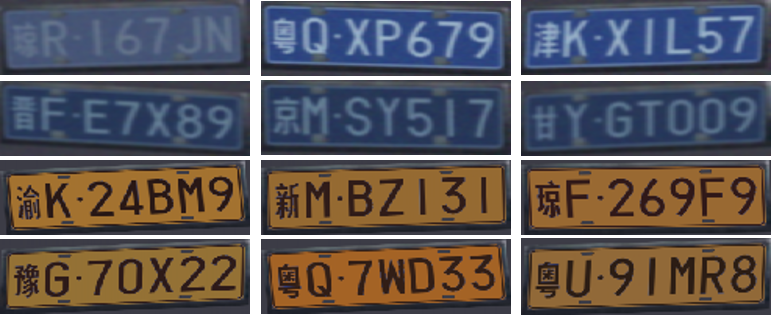}}
  \hspace{1in}
  \subfigure[CycleGAN Images]{
    \label{gan_imgs:b} %
    \includegraphics[width=2.5in]{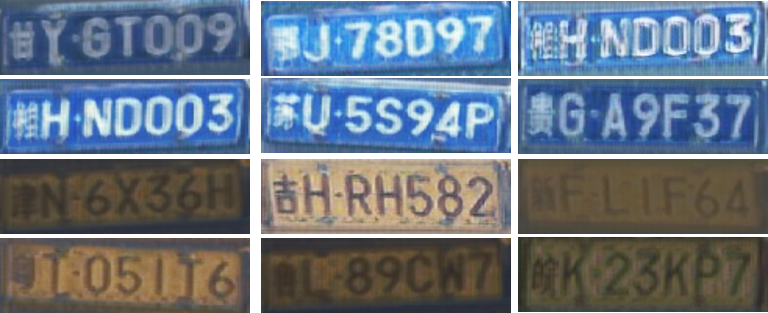}}

  \subfigure[CycleWGAN Images]{
    \label{gan_imgs:c} %
    \includegraphics[width=2.5in]{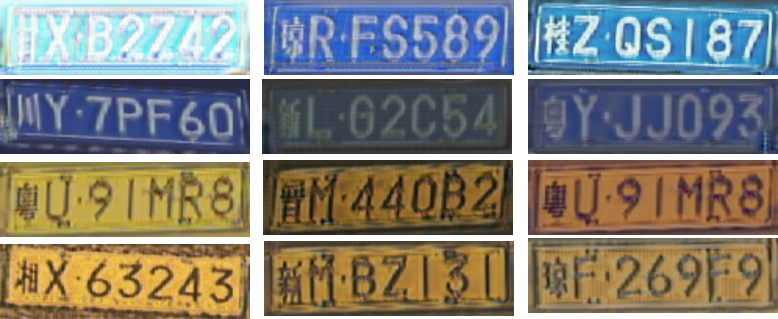}}
   \hspace{1in}
  \subfigure[Real Images]{
    \label{gan_imgs:d} %
    \includegraphics[width=2.5in]{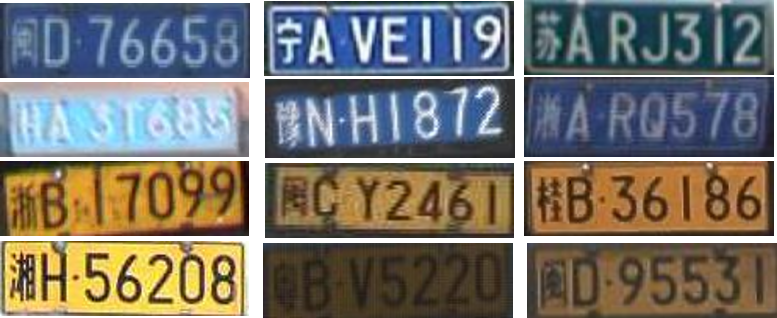}}

\caption{(a) The examples of license plates generated by our scripts (simple graphics with hand-crafted rules;
in other words, colors and character deformation are hard-coded).
    Note that these images are generated with labels. (b) The examples of license plates generated by CycleGAN \cite{zhu2017unpaired} model, using synthetic images as input. (c) The examples of license plates generated by CycleWGAN model, in which WGAN \cite{arjovsky2017wasserstein} techniques are applied. (d) Real license plates from Dataset-1.}
  \label{gan_imgs} %
\end{figure*}

\subsection{Implementation Details}
\textbf{The recognition framework} The convolutional RNN framework is similar to
that of  \cite{shi2016end} and \cite{li2016reading}, We implement it using Tensorflow \cite{abadi2016tensorflow}.
The images are first resized to $160 \times 48$ and then fed to the recognition framework.
We modify the last fully-connected layer to
have 68 neurons according to the 68 classes of label---31 Chinese characters, 26 letters, 10 digits and ``blank''.
Note that, the decoding method that we use in the above decoding process, independent of top-1 or top-$N$,
is based on a greedy
strategy. It means that we get the best path---most probable path, not the most probable labelling. Graves et al.\
\cite{graves2006connectionist} find the almost best labeling using prefix search decoding, which achieves a slight
improvement over best path decoding, at the cost of expensive computation and decoding time. Same as
\cite{shi2016end,li2016reading}, we adopt the best path decoding in inference.

\begin{figure}[htbp]
\centering
\includegraphics[width=2.5in]{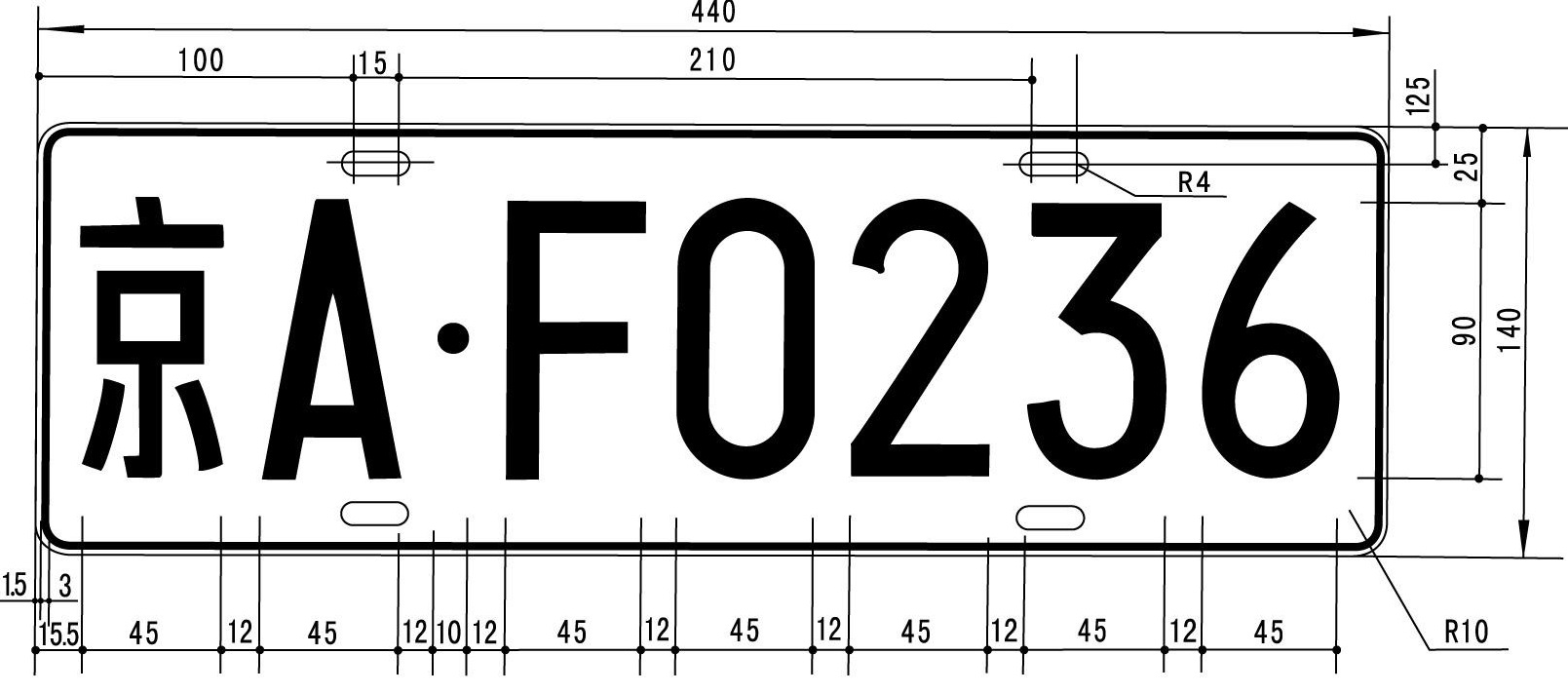}
\caption{Standard Chinese license plate. The first character is Chinese character representing region. The second must be a letter. And the last 5 characters can be letter (exclude ``I'' and ``O'') or digit, but there should be no more than two letters.}
\label{standardLP}
\end{figure}

\textbf{Synthetic data}
    We generate
    200,000 synthetic  license plates as our SynthDataset using computer graphics scripts,
    following the fonts, composition rules and colors, adding some Gaussian
    noise, motion blurring and affine deformation.
    The composition rules is according to the standard of Chinese license
    plates, such as the second character must be letter and at most two letters (exclude ``I'' and ``O'') in last five
    characters, as depicted in Fig. \ref{standardLP}. Some synthetic license plates are shown in Fig. \ref{gan_imgs:a}.

\textbf{GAN training and testing} We first train the CycleGAN model on 4,500 synthetic blue license plates from SynthDataset
and 4,500 real blue license plates from training set of Dataset-1, as depicted in Fig. \ref{pipeline}.
We follow the settings in \cite{zhu2017unpaired}.
The $\lambda$ in
Equation $(1)$ is set to 10 in our experiments.
All the images are scaled to $143 \times 143$, cropped to $128 \times
128$ and randomly flipped for data augmentation.
We use Adam with $\beta_1$ = 0.5, $\beta_2$ = 0.999 and learning rate of 0.0002. We save the
model for each epoch and stop training after 30 epochs. When testing, we use the last 20 models to generate 100,000 blue
license plates. The same goes for yellow license plates. Finally, we obtain 200,000 license plates that acquire the
license number and realistic style from SynthDataset and Dataset-1 separately. When applying the training method of
WGAN, $d_{iter}$ is set to 5, and the parameters of discriminator are clamped to $[-0.01, 0.01]$ after each gradient
update. \textit{Note that only the training set of Dataset-1 is used when training the GAN models.} For
data augmentation, 200,000 images are
randomly cropped to obtain 800,000 images before fed to the convolutional RNN models.

\begin{table}[t]
\footnotesize
\centering
\caption{Comparison of models using different frameworks on Dataset-1. Recognition accuracy (\%) and character recognition accuracy (\%) accuracy are listed.}
\label{cnn_compare}
\begin{tabular}{ccc}
\hline
& RA (top-1) & CRA  \\
\hline
SVM + ANN \cite{EasyPR} & 68.2& 82.5 \\
CNN + LSTMs  & 96.1 & 98.9 \\
\hline
\end{tabular}
\end{table}

\begin{table}[ht]
\footnotesize
\centering
\caption{Comparison of models that use different generated images for supervised learning without real images. The performance is evaluated on test set of Dataset-1. ``Random''  represents the randomly initialized model. Recognition accuracy (\%) and character recognition accuracy (\%) are listed. ``CRA-C'' is the recognition accuracy (\%) of Chinese character which is the first character, and ``CRA-NC'' is the recognition accuracy (\%) of letters and digits which are the last six characters.}
\label{onlyfake_Dataset_1}
\begin{tabular}{ccccc}
\hline
& RA (top-1) & CRA & CRA-C & CRA-NC\\
\hline
Random    & 0.0  & 0.8 & 0.0 & 0.9 \\
Script 	  & 4.4  & 30.0 & 20.0 & 31.7\\
CycleGAN  &	34.6 & 82.8 & 41.3 & 89.8 \\
CycleWGAN & \textbf{61.3} & \textbf{90.6}	& \textbf{66.2}& \textbf{94.8}\\

\hline
\end{tabular}
\end{table}

\begin{table}[t]
\footnotesize
\centering
\caption{Comparison of using different volumes of the real data and synthetic data generated by different methods on Dataset-1, i.e., scripts and CycleWGAN model. It means the generated images are used for training first, then specific  proportion of Dataset-1 is fed to the model. The performance is evaluated on the test set of Dataset-1. The CRNN baseline that trained on Dataset-1 only is also provided. Recognition accuracy (\%), character recognition accuracy (\%), top-3 recognition accuracy (\%) and top-5 recognition (\%) accuracy are listed.  }
\label{Dataset_1_20w}

\begin{tabular}{cccccc}
\hline
Training Data & Methods & RA (top-1) & CRA & top-3 & top-5 \\
\hline
\multirow{3}{*}{ 9$k$ } &
Baseline & 86.1 & 94.9 & 90.1 & 92.3 \\
& Script 	 & 90.2 & 97.0 & 94.6 & 95.6 \\
& CycleWGAN 	 & \textbf{93.6} & \textbf{98.4} & \textbf{96.8} & \textbf{97.4} \\
\hline
\multirow{3}{*}{ 50$k$ } &
Baseline & 93.1 & 97.9 & 96.4 & 97.2 \\
& Script 	 & 95.2 & 98.8 & 97.7 & 98.1 \\
& CycleWGAN 	 & \textbf{96.3} & \textbf{99.2} & \textbf{98.3} & \textbf{98.8} \\
\hline
\multirow{3}{*}{ 200$k$ (All) } &
Baseline & 96.1 & 98.9 & 98.0 & 98.5 \\
& Script	 & 96.7 & 99.1 & 98.6 & 98.8 \\
& CycleWGAN 	 & \textbf{97.6} & \textbf{99.5} & \textbf{98.9} & \textbf{99.2} \\
\hline
\end{tabular}
\end{table}

\begin{table}[ht]
\footnotesize
\centering
\caption{Comparison of the baseline on Dataset-2. We fine-tune the model on Dataset-2 after pretraining on CycleWGAN images. Recognition accuracy (\%), character recognition accuracy (\%), top-3 recognition accuracy (\%) and top-5 recognition (\%) accuracy are shown.}
\label{Dataset_2}

\begin{tabular}{ccccc}
\hline
 & RA (top-1) & CRA & top-3 & top-5 \\
\hline
Baseline & 94.5 & 98.4 & 97.6 & 98.1 \\
CycleWGAN 	 & \textbf{96.2} & \textbf{99.2} & \textbf{98.7} & \textbf{99.1} \\
\hline
\end{tabular}
\end{table}

\subsection{Evaluation}
\textbf{The convolutional RNN baseline model}
        We train the baseline models on two datasets directly.
        As shown in Table \ref{Dataset_1_20w}, we achieve a top-1 recognition accuracy
        of 96.1\% and character accuracy of 98.9\% on Dataset-1, without additional data. This baseline already yields a
        high accuracy, especially the top-3 and top-5 accuracy which are 98.0\% and 98.5\%. On Dataset-2, the
        recognition accuracy and character recognition accuracy are 94.5\% and 98.4\%, as shown in Table
        \ref{Dataset_2}.

        For comparison and to demonstrate the effectiveness of the CRNN framework on the task of license plate recognition, we
        also evaluate the performance of EasyPR \cite{EasyPR} on test set of Dataset-1, which is a popular open source
        software using support vector machines to locate plates and a single hidden layer neural network
        for character-level recognition. EasyPR is
        a segmentation based method.
        As shown in
        Table \ref{cnn_compare}, our approach yield a superior accuracy.

\textbf{Synthetic data only}
To directly evaluate the effectiveness of the data generated by different methods, we train the models using synthetic data generated by script, CycleGAN and CycleWGAN respectively. Then accuracy evaluations are performed, which are listed in Table \ref{onlyfake_Dataset_1}.

When we only use the SynthDataset which is generated by script for training, the recognition accuracy on test set of Dataset-1 is just 4.4\%.
To rule out the possibility of random initialization, we take a test on the model with random initial weights, which
produces
a result of 0.0\% and 0.8\% on recognition accuracy and character recognition accuracy.
It means these synthetic images generated by script do help.
However, compared with real license plates in Fig. \ref{samples_Dataset_1}, our synthetic license plates generated by script are overly simple.

The CycleGAN images achieve a recognition accuracy of 34.6\%.
The improvement of recognition accuracy from 4.4\% to 34.6\% demonstrates that the distribution generated by CycleGAN is much closer to the real distribution, which means that the generator performs a good mapping from synthetic data to real data with the regularization of discriminator. Some of the samples are shown in Fig. \ref{gan_imgs:b}. However, we observe the phenomenon that the outputs of generator tend to converge to a specific style, which shows some kind of mode collapse \cite{goodfellow2014generative,radford2015unsupervised,im2016generating}, the well-known problem of GAN training.

As shown in Fig. \ref{gan_imgs:c}, the CycleWGAN images show more various styles of texture and colors. Part of them are really undistinguishable from real images.
Although to the naked eyes the images’ quality may be slightly lower, which is similar to the observation in \cite{arjovsky2017wasserstein}, CycleWGAN shows aggressive rendering ability that prevents from mode collapse.
When we use the CycleWGAN images, we yield 61.3 RA, which indicates the CycleWGAN images make a relatively improvements of 1300\% compared to the script images.
Note that, the recognition accuracy of non-Chinese characters (last six characters) has already achieved 94.8\% without using any real data, as letters and digits are much easier than Chinese characters.

The results are consistent with the appearance of the images in Fig. \ref{gan_imgs}. We conclude that the trained model is more accurate if the synthetic data is more real.

{\bf Curriculum learning}
To further explore these synthetic images' significance, we then fine tune the pre-trained models on Dataset-1.
Thus,
we follow a curriculum learning strategy to train the system
with gradually more complex training data. As in the
first step, the large number of synthetic images
whose appearance is simpler than real images, and a
large lexicon of license plate numbers. At the same time,
the first step finds a good initialization of the appearance
model for the second step which uses real images.

As shown in Table \ref{Dataset_1_20w}, with the help of CycleWGAN images, we obtain
an impressive improvement of 1.5 pp over such a strong baseline.

To provide more general insights about the ability of GAN to generate synthetic training data, we adopt a cross-dataset strategy where the CycleWGAN is trained on data from Dataset-1 and produces images that will be used to train a model on Dataset-2.
Thus, the model pre-trained on CycleWGAN images is fine tuned on Dataset-2, with the same procedure. We observe an improvement of 1.7 pp (from 94.5\% to 96.2\%) in Table \ref{Dataset_2}. The experimental results on the Dataset-1 and Dataset-2 demonstrate the effectiveness: even on a strong baseline, these CycleWGAN images effectively yield remarkable improvements.

\textbf{Smaller datasets, more significant improvements}
		What if we only have a small real dataset? It is a common problem faced by some special tasks,
        especially when related to personal privacy.

        We  sample a small set of 9,000 ground-truth labeled license plate images from Dataset-1,
        For the first method, we train our recognition networks directly using these 9,000 license plates. The second
        method is the pipeline proposed in Section 1 that we use
        this dataset to train a GAN model and generate synthetic license plates.
        We then fine-tune this pre-trained model using 9,000 real
        images.

        As shown in Table \ref{Dataset_1_20w}, when we add 800$k$ GAN images to the network's training, our
        method significantly improves the LPR performance. We observe an impressive improvement of 7.5 pp (from
        86.1\% to 93.6\%). When using 50,000 real images for the fine-tuning instead of 9,000, we also observe impactful
        improvement of 3.2 pp (from 93.1\% to 96.3\%). It is clear that the
        improvement is more remarkable with a smaller training set of real images. What is more, the model trained on 9,000 training
        set and GAN images deliver better performance than the model trained on a training set with size of 50,000.
        This  comparison can also be found between 50$k$ and 200$k$ in Table \ref{Dataset_1_20w}.
        {\em It
        means that we can train a better model on a smaller real training set using this method,
        }
        validating the effectiveness of the proposed pipeline.

\begin{figure}[htbp]
\centering
\includegraphics[width=2.5in]{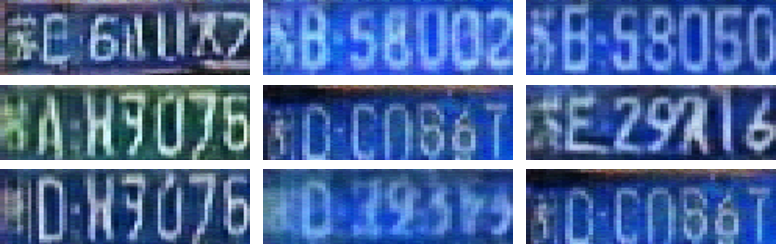}
\caption{License plate samples generated by a DCGAN model trained on part of Dataset-1. Through the all-in-one method, they are mixed with real license plates to regularize the CRNN model.}
\label{samples_dcgan}
\end{figure}

\begin{table}[ht]
\footnotesize
\centering
\caption{Comparison of mixing different amount of real images with DCGAN images. For all-in-one method, 50,000 DCGAN images is mixed with real images. Recognition accuracy (\%), character recognition accuracy (\%), top-3 recognition accuracy (\%) and top-5 recognition (\%) accuracy are shown.}
\label{unsp_Dataset_1}
\begin{tabular}{ccc}
\hline
& RA (top-1) & CRA \\
\hline
Dataset-1-9k & 86.1 & 94.9 \\
Dataset-1-9k + All in one 	 & 78.1 & 92.8 \\
\hline
Dataset-1-50k & 93.1 & 97.9 \\
Dataset-1-50k + All in one 	 & 92.6 & 97.7 \\
\hline
Dataset-1-200k & 96.1 & 98.9 \\
Dataset-1-200k + All in one 	 & 96.0 & 98.9 \\
\hline
\end{tabular}
\end{table}

\textbf{Unlabeled DCGAN images}
	To compare the effectiveness of labeded CycleWGAN images and unlabeled images generated by DCGAN \cite{radford2015unsupervised}, we train a DCGAN  model to generate unlabeled images, and  apply the all-in-one \cite{salimans2016improved,odena2016semi,zheng2017unlabeled} method to them for a semi-supervised learning. All-in-one is an approach that takes all generated unlabeled data as one class.\par
	DCGAN provides a stable architecture for training GANs and can only generate unlabeled images. It consists of one generator and one discriminator. We follow the settings in \cite{radford2015unsupervised}. For the generator, we use four fractionally-strided convolutions after projecting and reshaping a 100-dim random vectors to $4 \times 4 \times 16$ to generate images with size of $64 \times 64 \times 3$. For the discriminator, a fully-connected layer following four stride-2 convolutions with kernel size of $5 \times 5$ is used to perform a binary classification that whether the image is real or fake.
	We train the DCGAN model on 46,400 images from Dataset-1, and then use the model to generate 50,000 images by
    inputting 100-dim random vectors in which each entry is in a range of $[-1, 1]$.
    Some samples are shown in Fig. \ref{samples_dcgan}.

    For all of our license plates, there are 68 different characters, which means 68 classes. We create a new character class, and every character in generated license plate is assigned to this class. Same as \cite{zheng2017unlabeled}, these generated images and ground truth images are mixed and trained simultaneously. As shown in Table \ref{unsp_Dataset_1}, we explore the effect of adding these 50,000 images to Dataset-1-9k, Dataset-1-50k and Dataset-1-200k.
    The results show that the DCGAN images fail to obtain improvements, which is opposite to the results of \cite{zheng2017unlabeled}. We think it is because that the bias in the task of person re-identification does not exist in LPR.

\begin{figure}[htbp]
\centering
\includegraphics[width=2.5in]{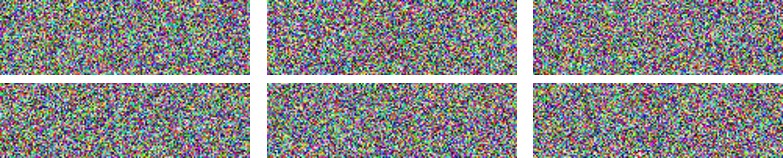}
\caption{Random images that consist of random value at each pixel. They are input to a trained model to produce the condition distribution which can be viewed as the knowledge of the model.}
\label{random_images}
\end{figure}

\subsection{Interpretation}

Thus far we empirically prove the effectiveness of CycleGAN generated images for recognition.
However from a theoretical perspective,  it is not clear about the effect of images generated by GAN, in real learning tasks.
Thus, we perform some visualizations to interpret the effectiveness of GAN images. \par

The knowledge of a model can be viewed as the conditional distribution it produces over outputs given an
input \cite{hinton2015distilling,pereyra2017regularizing}.
Inspired by this insight, we input 5,000 random-generated images to models, then average the outputs of last
fully-connected layer together as a confidence map to simulate the output distribution.
Some samples of random images are shown in Fig. \ref{random_images}.
The confidence map namely $y$ is illustrated in Fig. \ref{output_map}, where $y^k_t$ can interpreted as probability of observing label $k$ at time $t$ as described in Section 3.2.

\begin{figure*}[ht]
	\centering
	\subfigure[800$k$ script images]{
		\label{vis_sp:script} %
		\includegraphics[width=0.34\textwidth]{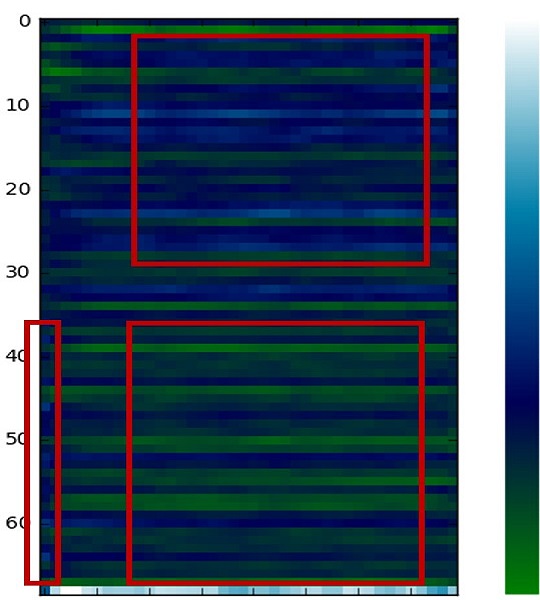}}
	\hspace{1in}
	\subfigure[800$k$ CycleWGAN images]{
		\label{vis_sp:gan} %
		\includegraphics[width=0.34\textwidth]{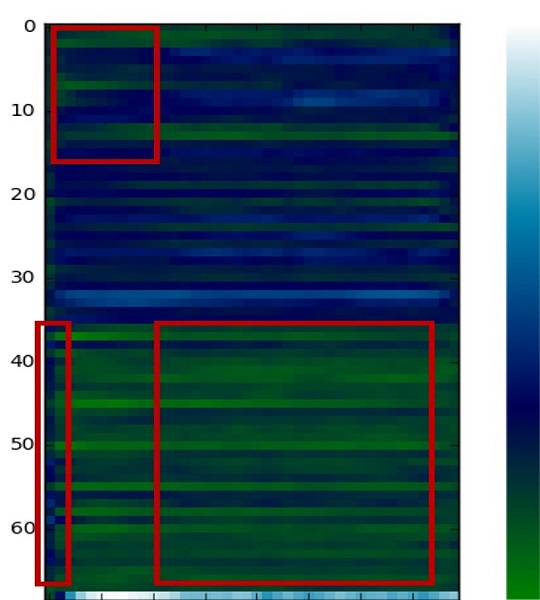}}
	
	\subfigure[200$k$ real after 800k CycleGAN images]{
		\label{vis_sp:gan_grtr} %
		\includegraphics[width=0.34\textwidth]{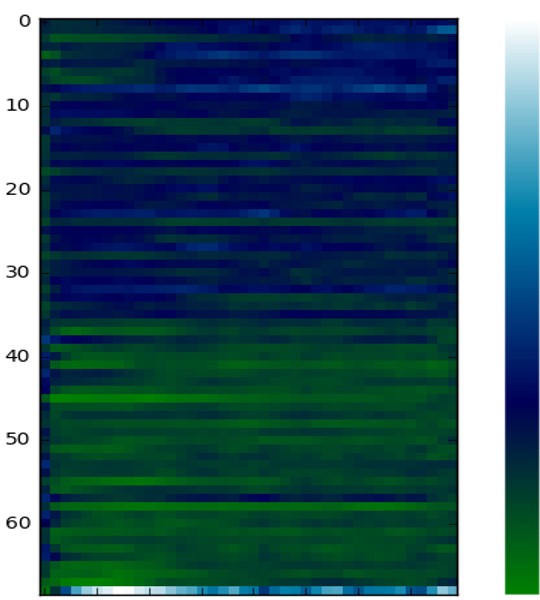}}
	\hspace{1in}
	\subfigure[200$k$ real images]{
		\label{vis_sp:grtr} %
		\includegraphics[width=0.34\textwidth]{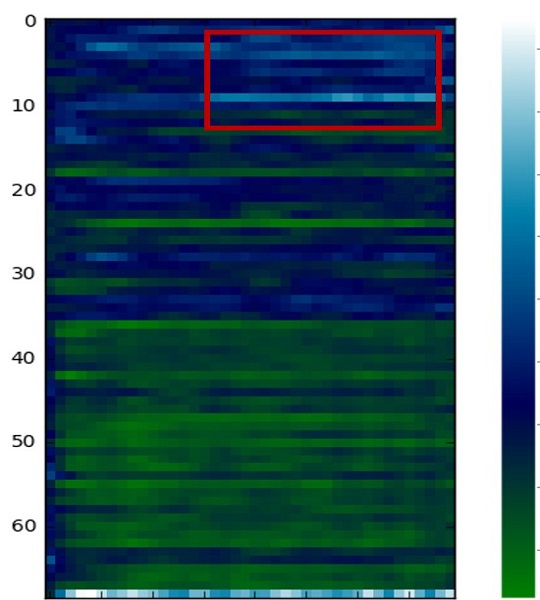}}
	\caption{Comparison of models trained on different datasets. The recognition probabilities on 68 classes are shown vertically (with classes order from top to
		bottom: 0-9, A-Z, Chinese characters and ``blank''). Confidence increases from green to white. (a) The model trained on
		800k script-generated images only. The blue area indicates a higher probability to be digits and letters. (b) The
		model trained on 800$k$ CycleWGAN images only. Here shows a more clear color distribution which indicates the composition rule of the Chinese license plates.
		(c) The model pre-trained on 800$k$ CycleWGAN images and fine-tuned on 200k real images.
		(d) The model trained on 200$k$ real images only. The high confidence in the red bounding box represent the not discriminative features. }
	\label{vis_comparison_sp} %
\end{figure*}

Here, we mainly explain this question: why do those labeled GAN images yield further improvements?
First, the model trained on script-generated images only is visualized, as shown in Fig. \ref{vis_sp:script}. We observe that the bottom of the first column is blue while the top is green, which means that it is more probable to observe Chinese characters in the place of license plate's first character. On the contrary, the rest places tend to be digits and letters.

When we observe the model trained on CycleWGAN images only, the composition rules of license plates are more clear,
as illustrated in Fig. \ref{vis_sp:gan}.
Besides, the columns next to the first column show concentrated probabilities in middle position, which indicate the second character tends to be letters. These observations match the ground truth rules explicitly.
The above results indicate the knowledge of composition rules that are hard-coded in scripts is learned by the model and
help in sequence labelling.

Once the model pre-trained on CycleWGAN images is fine-tuned on real images instead of script images, these general rules are still clear. However, the model trained on real images directly learns the non-general features, e.g., very high probability of ``9'' which is caused by the many ``9''s in training set, as depicted in Fig. \ref{vis_sp:grtr}.
The non-general features cause over-fitting.
Besides, we pick  the images that make up the improvements. About 35\% of them are caused by the first character (Chinese character), and about one quarter of them are caused by the confusing pairs like ``0'' \& ``D'', ``8'' \& ``B'' and ``5'' \& ``S''. It means that these CycleWGAN images help extract discriminative features of license plates.

In a word, the knowledge and prior hard-coded in script does help; the GAN images soften these knowledge and make output distribution be
close to ground-truth distribution which lead to further improvements.
It is consistent with the experimental results in
Section 4.4. When training data is large, the knowledge in additional GAN images help less because the real training set
is sufficiently large  to carry most of these knowledge.

\subsection{Model Compression}
To further meet requirements of inference on mobile and embedded devices, we replace the standard convolution with
depth-wise separable convolution \cite{sifre2014rigid} and adopt the hyper-parameter width multiplier proposed by
\cite{howard2017mobilenets}. A depth-wise separate convolution contains a depth-wise convolution and a 1$\times$1
(point-wise) convolution. For depth-wise convolution, a single filter is applied to each input channel separately. Then a pointwise convolution is applied which is a 1$\times$1 convolution that combines different channels to create new features. This factorized convolution efficiently reduces the model size and computation.

With stride one and padding, a standard convolution filtering on a $H \times W \times M$ feature map $R$ outputs a $ H \times W \times N $ feature map O. Here, $H$ and $W$ represent the height and width of the feature map. $M$ and $N$ are the depth of the input and output feature map. While the size of convolution kernel $K$ is $F \times F \times M \times N$ where $F$ denotes the spatial dimension of the kernel, we can compute the output feature map $O$ as:
\begin{equation}
O_{p,q,n} = \sum_{i,j,m}K_{i,j,m,n} \cdot R_{p + i - 1, q + j - 1, m}.
\end{equation}
During this forward computation, the computation cost is defined as below:
\begin{equation}
C_{\rm standard} = F \cdot F \cdot M \cdot N \cdot H \cdot W.
\end{equation}
As for the depth-wise convolution, the temporary output feature map is computed as:
\begin{equation}
\hat{O}_{p,q,m} = \sum_{i,j} \hat{K}_{i,j,m} \cdot R_{p + i - 1, q + j - 1, m}.
\end{equation}
Then we perform a $1\times1$ convolution to produce the final output feature map of this depth-wise separate convolution.
Over all, the computation cost of the depth-wise separate convolution is:
\begin{equation}
F \cdot F \cdot M \cdot H \cdot W + M \cdot N \cdot H \cdot W.
\end{equation}

To explore the trade off between the accuracy and inference speed, we bring in a hyper-parameter called width
multiplier. At each layer, we multiply both input depth $M$ and output depth $N$ by $\alpha$. Then the computation
cost of the separate depth-wise convolution becomes:
\begin{equation}
C_{\rm separate} = F \cdot F \cdot \alpha M \cdot H \cdot W + \alpha M \cdot \alpha N \cdot H \cdot W.
\end{equation}
Dividing the $C_{\rm separate}$ by $C_{\rm standard}$, we then get the ratio of computation cost:
\begin{equation}
\begin{split}
\frac{C_{\rm separate}}{C_{\rm standard}} &= \frac{F \cdot F \cdot \alpha M \cdot H \cdot W + \alpha M \cdot \alpha N \cdot H \cdot W}{F \cdot F \cdot M \cdot N \cdot H \cdot W} \\
&= \alpha \frac{1}{N} + \alpha^2 \frac{1}{F^2}.
\end{split}
\end{equation}

If we use the default width multiplier of 1.0 and $3\times3$ depth-wise separable convolutions, the computation cost becomes about 9 times less than the original standard convolution, while the output depth $N$ is large.

Based on this efficient depth-wise separable convolution and the existing CRNN framework, we construct a lightweight convolutional recurrent neural network and denote it as ``LightCRNN''. Note that only the convolutions are modified, while the parameters and computation cost in LSTM and the fully-connected layer in it are kept. The entire architecture is defined in Table \ref{lcrnn_config}.

\begin{table}[ht]
\footnotesize
\centering
\caption{Configuration of the LightCRNN model. ``k'', ``s'' and ``p'' represent kernel size, stride and padding size. Each convolution is followed by batch normalization and ReLU nonlinearities.}
\label{lcrnn_config}
\begin{tabular}{c|c}
\hline
\hline
Layer Type & Configurations \\
\hline
\hline
Bidirectional-LSTM & \#hidden units: 256\\
\hline
Bidirectional-LSTM & \#hidden units: 256\\
\hline

Conv & \#filters:512, k:$1 \times 1$, s:1, p:0 \\
\hline
Depthwise Conv &  k:$2 \times 2$, s:1, p:0 \\
\hline

Conv & \#filters:512, k:$1 \times 1$, s:1, p:0 \\
\hline
Depthwise Conv &  k:$2 \times 2$, s:1, p:0 \\
\hline

MaxPooling & s:$1 \times 2$, p:$1 \times 0$ \\
\hline
Conv & \#filters:512, k:$1 \times 1$, s:1, p:0 \\
\hline
Depthwise Conv &  k:$3 \times 3$, s:1, p:1 \\
\hline

MaxPooling & s:$1 \times 2$, p:$1 \times 0$ \\
\hline
Conv & \#filters:256, k:$1 \times 1$, s:1, p:0 \\
\hline
Depthwise Conv &  k:$3 \times 3$, s:2, p:1 \\
\hline

Conv & \#filters:128, k:$1 \times 1$, s:1, p:0 \\
\hline
Depthwise Conv &  k:$3 \times 3$, s:2, p:1 \\
\hline

Conv & \#filters:64, k:$3 \times 3$, s:1, p:1 \\
\hline

Input & $160 \times 48 \times 3$ RGB images\\
\hline
\hline
\end{tabular}
\end{table}

\begin{table}[ht]
\footnotesize
\centering
\caption{Comparison of using different models. All are trained on Dataset-1 and tested on its test set. Recognition accuracy (\%), model size (MB) and inference speed (FPS) are listed. The proposed architecture significantly decreases the model size and speeds up the inference.}
\label{lightcrnn_compare}
\begin{tabular}{cccc}
\hline
Model & RA (top-1) & Size & Speed\\
\hline
LightCRNN 	  & 96.5  & 40.3  & 13.9 \\
1.2 LightCRNN  & 97.0 & 44.2 & 11.5 \\
1.2 LightCRNN + CycleWGAN images & 98.6 & 44.2 & 11.5 \\
\hline
\end{tabular}
\end{table}

	We evaluate the proposed LightCRNN on Dataset-1 without GPU. The following experiments are carried out on a workstation with 2.40 GHz Intel(R) Xeon(R) E5-2620 CPU and 64GB RAM. The LightCRNN models are trained and tested using TensorFlow \cite{abadi2016tensorflow}. \textit{Only a single core is used when we perform forward inference}. As shown in Table \ref{lightcrnn_compare}, the LightCRNN not only efficiently decreases the model size from 71.4$MB$ to 40.3$MB$ and speeds up the inference from 7.2$FPS$ to 13.9$FPS$,
	but also increase the recognition accuracy.
	It is probably because this LightCRNN reduce the overfitting of the model.
	To adjust the trade off between latency and accuracy, we apply the width multiplier of 1.2. This 1.2 LightCRNN brings another improvement of 0.5 pp.
	We then apply the proposed pipeline which leverages CycleWGAN image data and yields an improvement of 1.6 pp.

	Thus, the combination of our proposed pipeline and architecture make it more close to perform accurate recognition on mobile and embedded devices.

\begin{figure}[htbp]
\centering
\includegraphics[width=2.5in]{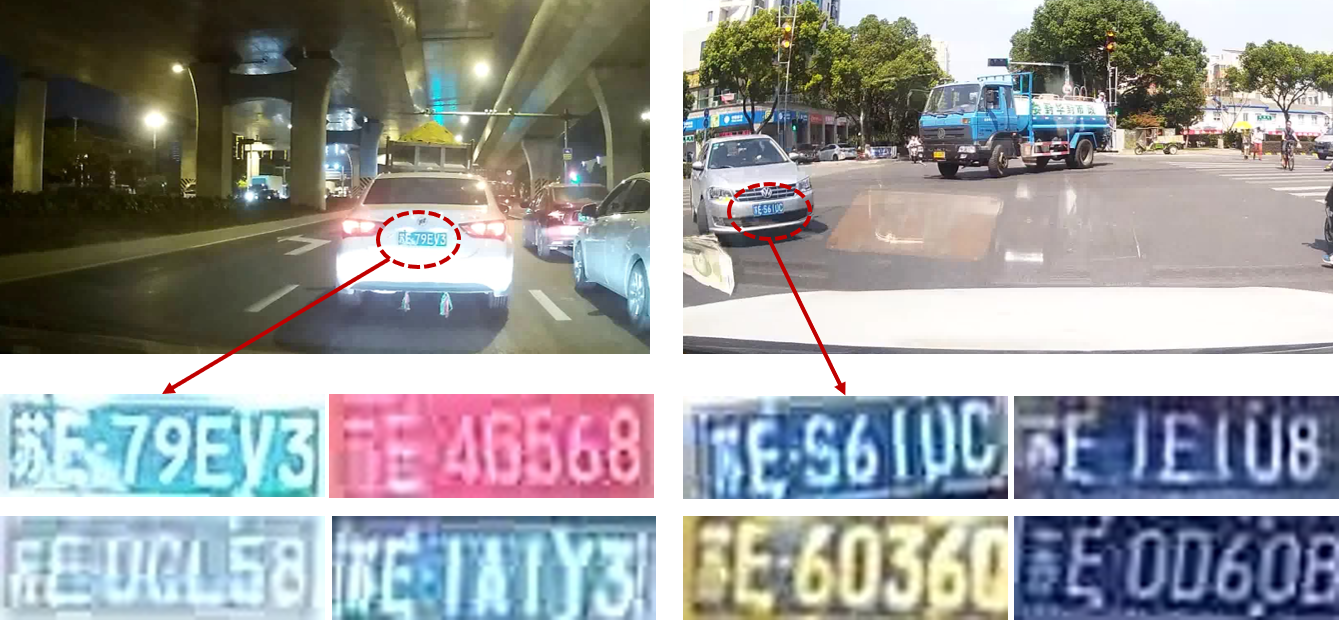}
\caption{Some samples of Dataset-3. The videos are shot when both the source and target objects are moving.}
\label{Dataset-3}
\end{figure}

\begin{figure}[htbp]
\centering
\includegraphics[width=2.5in]{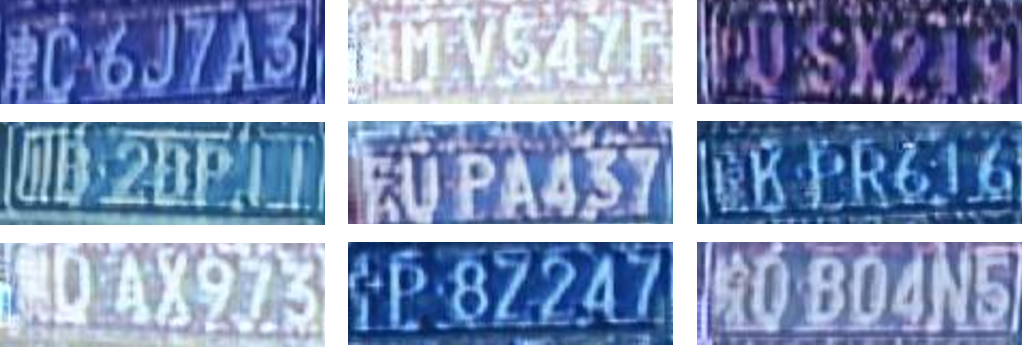}
\caption{Some samples of CycleWGAN images for moving LPR. The CycleWGAN models are trained using script images and training set of Dataset-3.}
\label{moving_gan_img}
\end{figure}

\begin{table}[ht]
\footnotesize
\centering
\caption{Results on test set of Dataset-3. Fine-tuning on the model pre-trained on the CycleWGAN images improves the baseline. Recognition accuracy (\%) and character recognition accuracy (\%) accuracy are listed.}
\label{Dataset-3-results}
\begin{tabular}{ccc}
\hline
& RA (top-1) & CRA  \\
\hline
Baseline  & 89.4 & 97.6 \\
CycleWGAN +  & \textbf{92.1} & \textbf{98.0} \\
\hline
\end{tabular}
\end{table}

\subsection{Moving LPR}
We further validate our pipeline on more complicated data.
We shoot in a moving car and capture the license plates of other vehicles. Both the viewpoint and target object are moving fast and irregularly. This data is much more hard than the normal  monitoring data which is collected at settled location and angle.
The LPR task then becomes more challenging and we denote it as the task of \textit{moving LPR}, which is with wide applications in driving assistance, surveillance and mobile robotics, etc.

The moving LPR has two main bottlenecks.
First, this harder task needs more examples and large-scale data are expensive to collected, as you can not just export the data from monitoring camera but have to shoot videos in various regions to meet the requirement of diversity.
Second, the models need to be more efficient when you deploy the systems in vehicles or drones.
Our proposed pipeline eases up the first bottleneck, and the proposed LightCRNN helps the latter.

These images are collected in Suzhou, a normal Chinese city.
They cover various situatiions, such as night, highway, crossroad and so on.
Some samples are shown in Fig. \ref{Dataset-3}.
After the images are annotated and cropped, we obtain 22026 license plates, namely, the Dataset-3.  2000 of them are splitted as test set of Dataset-3.

Considering the trade off between accuracy and latency, we adopt the 1.2x LightCRNN in our following experiments.
We first train a model using the training set of Dataset-3 and evaluate on the corresponding test set. We obtain a recognition accuracy of 89.4\%.
Then the proposed pipeline is applied. First, we train the CycleWGAN using script images and the training set of Dataset-3. Both are 20$k$ images. Then we feed the script images to the trained generators and obtain 800$k$ synthetic CycleWGAN images. We show some samples in Fig. \ref{moving_gan_img}.
The recognition model is pre-trained on the CycleWGAN images and fine-tuned on the training set of Dataset-3.
As thus, we yield an improvement of 2.7 pp, as listed in Table \ref{Dataset-3-results}. It indicates that our methods work well on even more hard LPR scenario.

\section{Conclusion}
In this paper, we integrate the GAN images into supervised learning, i.e., license plate recognition.
Using our method, large-scale realistic license plates with arbitrary license numbers are generated by a CycleGAN model
equipped with WGAN techniques. The  convolutional RNN baseline network is trained on GAN images first, and then real
images in a curriculum learning manner.

Experimental results demonstrate that the proposed pipeline brings significant improvements on different license plates datasets. The significance of GAN-generated data is magnified when real annotated data is limited.
Furthermore, the question that why and when do these GAN images help in supervised learning is answered, using clear visualizations. We also propose a lightweight convolutional RNN
model using depth-wise separate convolution, to perform fast and efficient inference.

\appendices
\section*{Acknowledgment}
The work was supported by the Shanghai Natural Science Foundation (No. 17ZR1431500).

{\footnotesize
\bibliographystyle{ieee}
\bibliography{paper}
}

\ifCLASSOPTIONcaptionsoff
\newpage
\fi

\end{document}